\documentclass[journal]{IEEEtran}
\ifCLASSINFOpdf
\else
\fi
\usepackage{graphics} 
\usepackage{mathtools}
\usepackage{times} 
\usepackage{amsmath} 
\usepackage{amssymb}  
\usepackage{xcolor}
\usepackage[utf8]{inputenc}
\usepackage{comment}
\usepackage{algorithm,algorithmic}
\usepackage{amsthm}
\usepackage{balance}
\usepackage{hyperref}

\usepackage{multirow}
\usepackage{graphicx}
\usepackage{booktabs} 

\usepackage{tikz}

\DeclareMathAlphabet\mathbfcal{OMS}{cmsy}{b}{n}

\usepackage{cancel}
\theoremstyle{plain}

\newcommand{\q}{\boldsymbol{q}}
\newcommand{\R}{\mathbb{R}}

\newcommand{\x}{\boldsymbol{x}}

\DeclareMathAlphabet{\mathcal}{OMS}{cmsy}{m}{n}

\definecolor{myplank}{rgb}{ 0.8510 ,   0.6627  ,  0.4314}
\usepackage[dvipsnames]{xcolor}
\begin{document}
%


\title{Walk the \textcolor{Tan}{PLANC}: \underline{P}hysics‑Guided R\underline{L} for \underline{A}gile Humanoid Locomotio\underline{N} on \underline{C}onstrained Footholds}
%
%
%

\author{Min~Dai$^{1}$,
        William~D.~Compton$^{2}$,
        Junheng~Li$^{1}$,
        Lizhi~Yang$^{1}$,
        and~Aaron~D.~Ames$^{1,2}$
\thanks{$^{1}$Authors are with the Department
of Mechanical and Civil Engineering, California Institute of Technology,
Pasadena, CA 91125.}      
\thanks{$^{2}$Authors are with the Department
of Computing and Mathematical Sciences, California Institute of Technology,
Pasadena, CA 91125.}%
\thanks{This research is supported by Technology Innovation Institute (TII).}
\\ \vspace{0.2cm} 
\textit{Project Website: \url{https://caltech-amber.github.io/planc/}}
\vspace{-0.9cm}

}

\maketitle


\begin{abstract}
Bipedal humanoid robots must precisely coordinate balance, timing, and contact decisions when locomoting on constrained footholds such as stepping stones, beams, and planks---even minor errors can lead to catastrophic failure. Classical optimization and control pipelines handle these constraints well but depend on highly accurate mathematical representations of terrain geometry, making them prone to error when perception is noisy or incomplete. Meanwhile, reinforcement learning has shown strong resilience to disturbances and modeling errors, yet end-to-end policies rarely discover the precise foothold placement and step sequencing required for discontinuous terrain. These contrasting limitations motivate approaches that guide learning with physics-based structure rather than relying purely on reward shaping. In this work, we introduce a locomotion framework in which a reduced-order stepping planner supplies dynamically consistent motion targets that steer the RL training process via Control Lyapunov Function (CLF) rewards. This combination of structured footstep planning and data-driven adaptation produces accurate, agile, and hardware-validated stepping-stone locomotion on a humanoid robot, substantially improving reliability compared to conventional model-free reinforcement-learning baselines. The entire open-sourced code base is available at \url{https://anonymous.4open.science/r/robot_rl-E4FF}, and the supplementary video is provided at \url{https://youtu.be/D_CZ_FRYjtc}.
\end{abstract}


%
\IEEEpeerreviewmaketitle

\section{Introduction}

Achieving robust and adaptive bipedal locomotion over discrete footholds such as stepping stones remains a fundamental challenge in legged robotics. These conditions are critically relevant for humanoid robots deployed in disaster zones or cluttered industrial environments, where viable contact regions are sparse, discontinuous, and bounded by strict safety margins. Unlike traversing continuous rough terrain, where small foot placement errors can often be tolerated, stepping-stone environments impose hard constraints on contact location and timing. A single misstep or localized slip can lead to unrecoverable failure, making full-body motion coordination significantly more challenging.

Traditionally, terrain-aware locomotion has been approached through model-based control strategies that prioritize physical consistency and safety constraints. 
In one school of thought, the challenge of adapting to discrete terrain is delegated to a dedicated footstep planner with quasi-static walking, often built on reduced-order models (RoMs) such as the Linear Inverted Pendulum (LIP)~\cite{ griffin_footstep_2019} and centroidal dynamics \cite{ponton_efficient_2021, kojio_footstep_2020}. Within this hierarchical structure, terrain reasoning can be explicitly encoded as a mixed-integer optimization problem with explicit kinematic and foothold constraints \cite{deits_footstep_2014}, or solved through sampling-based methods for safe and efficient foot-step generation \cite{perrin_fast_2012}.
Instead of optimizing over the footstep locations,
offline gait libraries generated with full order models \cite{grizzle_models_2014} can be used with control barrier functions (CBFs) to modulate foot placement during execution~\cite{nguyen_3d_2016,nguyen_dynamic_2018}. 
More recently, researchers have begun to unify footstep planning and locomotion control within a single optimization framework, enabling simultaneous reasoning over control actions and footstep locations \cite{dai_bipedal_2022, acosta_bipedal_2023, xiang_adaptive_2024}. While promising, these non-convex formulations impose substantial computational demands and remain challenging on high–DoF humanoid robots over discrete terrains.
Consequently, many valid approaches are restricted to quasi-static motions  \cite{fallon_continuous_2015}. 

\begin{figure}
    \centering
    \includegraphics[width=\linewidth]{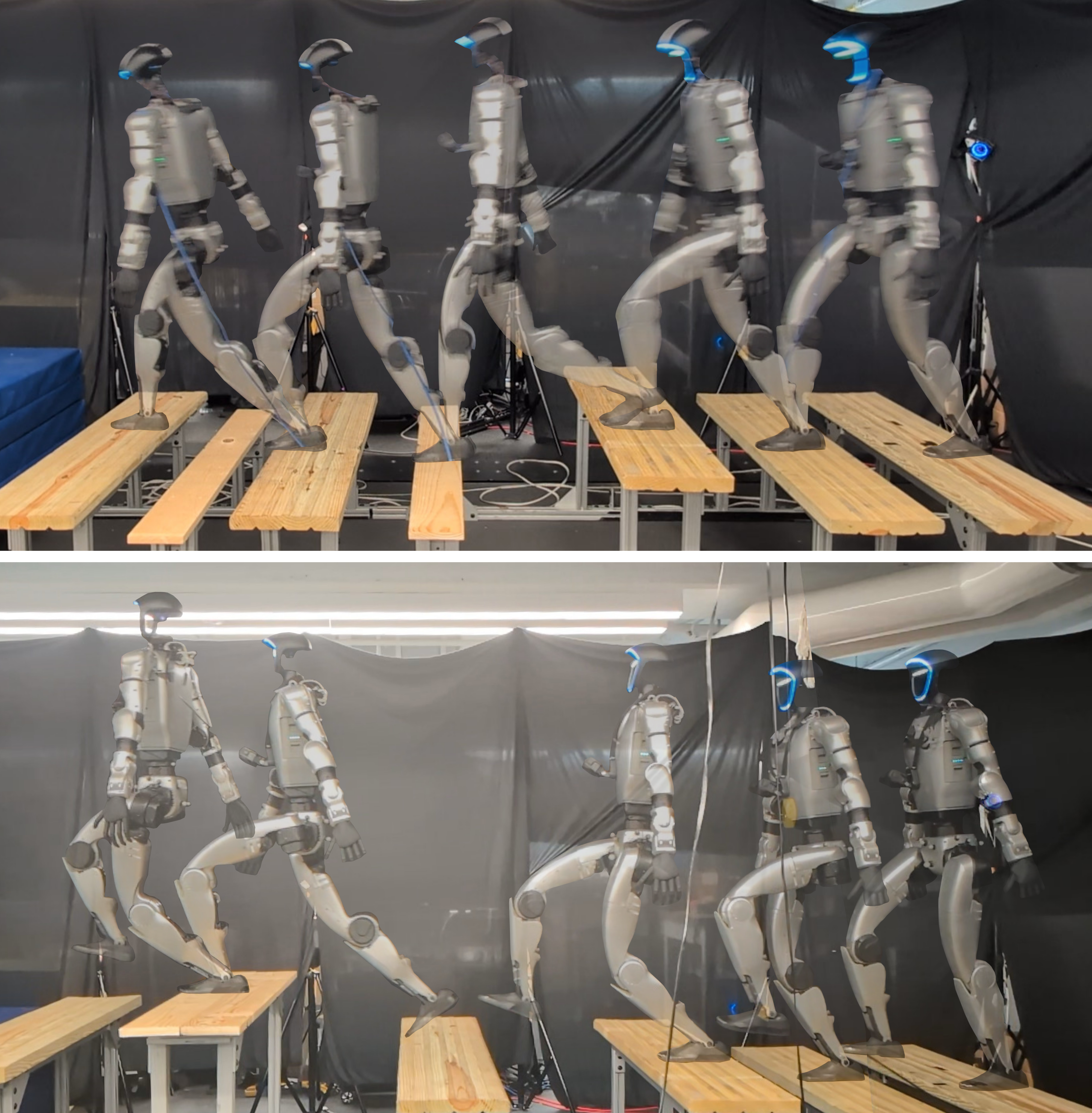} 
    \vspace{-5mm}
    \caption{Model-guided RL traversing constrained footholds on the Unitree G1 humanoid robot. The policy is trained with environmentally consistent references generated dynamically during training by a momentum-regulated linear inverted pendulum for precise footstepping and center of mass regulation, and is deployed successfully on hardware.}
    \label{fig:hero}
    \vspace{-5mm}
\end{figure}

\begin{figure*}[!t]
    \centering
    \includegraphics[width=\linewidth]{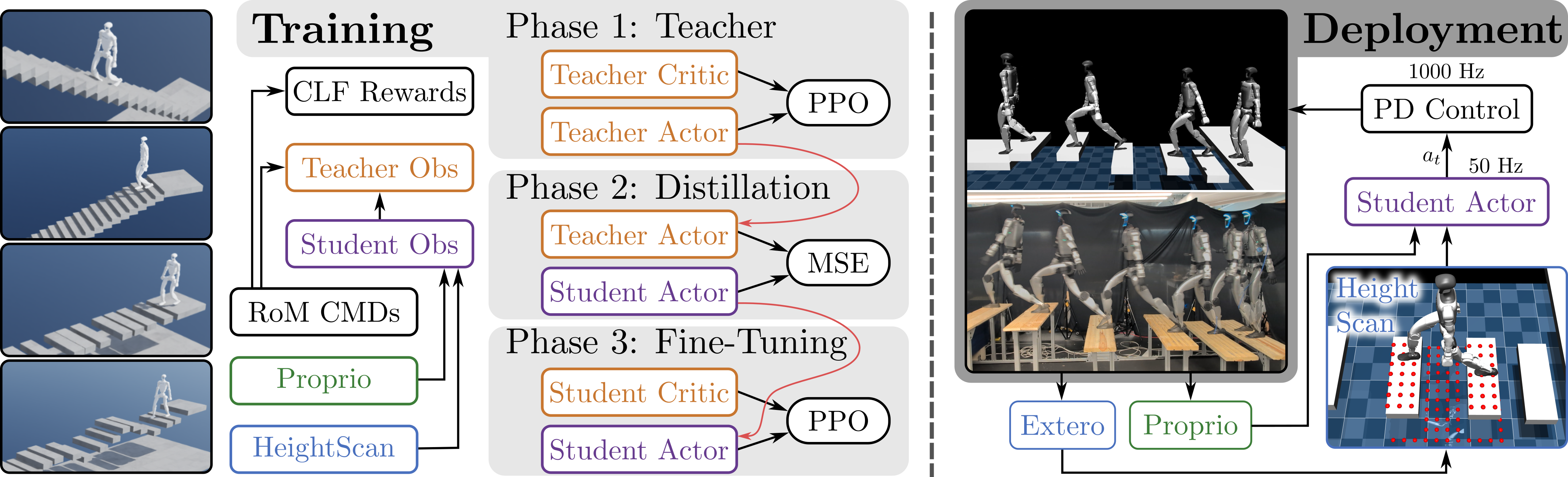} 
    \vspace{-5mm}
    \caption{A visual depiction of the model-guided RL architecture used to achieve stepping stones. The left column shows the four terrains trained in simulation, (top to bottom) stairs up, stairs down, flat stones, and height varying stones. Training is carried out in a teach-student fashion, with CLF rewards informed by the RoM. The student policy is successfully deploy in sim-to-sim transfer and on hardware.}
    \label{fig:arch_diagram}
    \vspace{-5mm}
\end{figure*}

Furthermore, a primary bottleneck for model-based methods lies in the integration of perception. The terrain dependency necessitates a rigorous terrain segmentation and state estimation pipeline~\cite{acosta_perceptive_2025}. 
In practice, purely model-based locomotion pipelines are already fragile, as they rely on highly accurate modeling and state feedback. Extracting precise mathematical representations for terrain geometry and foothold representations from noisy point clouds adds additional latency, uncertainty, complexity, and conservatism, as even minor segmentation or odometry errors can cause the planner to impose invalid constraints \cite{Fallon2019PlaneSeg}. This dependence on clean, structured sensor data places unreasonably heavy demands on the perception module and makes it difficult to deploy pure model-based approaches in unstructured environments.

Conversely, deep reinforcement learning (RL) has emerged as a powerful alternative, demonstrating the ability to generate robust locomotion directly from high-dimensional sensor data, including height maps. RL policies naturally integrate proprioception and exteroception, learning robust behaviors that generalize to hardware deployment. However, standard RL pipelines often prioritize survival over precision. While successful on continuous rough terrain \cite{cui_adapting_2024, gu_advancing_2024}, these methods struggle with the sparse foothold task, where precise end-effector control is paramount.

Early attempts to bridge this gap incorporated ground-truth terrain information directly into the RL observation space to guide foot placement \cite{xie_allsteps_2020}, thus limited to simulation with perfect step knowledge. \cite{singh_learning_2022} focused on structured environments like stairs, but demonstrated limited foot placement accuracy and lacked real-world validation. More recently, vision-based RL has achieved hardware deployment \cite{duan_learning_2022, duan_sim--real_2022, kim_learning_2025}, but these policies frequently exhibit low success rates in sparse-contact scenarios. Furthermore, unconstrained end-to-end approaches \cite{wang_beamdojo_2025, he_attention-based_2025} often exploit reward structures to maximize stability at the cost of motion quality, resulting in unnatural or period-two gaits. These methods can also be computationally prohibitive, with architectures like~\cite{he_attention-based_2025} requiring days of training.

The dichotomy is clear: model-based control excels at planning feasible motion under strict constraints (timing and placement) but is brittle to uncertainty; RL excels at robust execution under uncertainty but struggles to discover precise, constrained motions from scratch. This suggests that relying solely on reward engineering to induce precise stepping is inefficient. 
There is a growing consensus toward guided RL, where structured human trajectory references \cite{liao_beyondmimic_2025, liu2025opt2skill} or model-based references \cite{suliman_reinforcement_2025, lee_integrating_2024,jenelten_dtc_2024, li_clf-rl_2025} constrain the learning manifold. By shaping the learning process around feasible reference motions, RL can focus on refining contact dynamics and disturbance rejection rather than rediscovering the fundamentals of walking.

The core contributions of this work are the following,
\begin{itemize}
    \item We propose a physics-guided reinforcement learning framework that tightly integrates a reduced-order, terrain-aware bipedal stepping-stone planner with CLF-based rewards design into the RL training loop, outlined in Fig. \ref{fig:arch_diagram}. The framework generates terrain-consistent references for constrained foot placement, CoM trajectory, and step timing for RL training with minimal computation overhead, enabling agile and safe locomotion over discrete aperiodic terrains.

    \item We remove the dependency and extended efforts on human motion dataset collections and retargeting pipelines in a multi-environment generalist locomotion policy. The model guidance provides a physically grounded, controllable, and transferable prior for RL in severely constrained environments.

    \item Our approach is both numerically and experimentally validated on the Unitree G1 humanoid robot over randomized stepping-stone terrains with gap and height variances, demonstrating improved stability, precise foot placement, generalization to unseen terrain, and consistent performance gains over model-free baselines.

\end{itemize}


\section{Preliminaries}
\label{sec:preliminaries}
\subsection{Bipedal Robot Hybrid Dynamics}
A bipedal robot can be modeled as a hybrid dynamical system
$\Sigma = \left( \mathcal{D}, \mathcal{U}, \mathcal{G}, \Delta, f, g \right)$,
where \( \mathcal{D} \) denotes the set of continuous domains, 
\( \mathcal{U} \subset \mathbb{R}^{n_u} \) is the control input space, 
\( \mathcal{G} \subset \mathcal{D} \) represents the guards corresponding to impact events, 
\( \Delta: \mathcal{G} \to \mathcal{D} \) defines the discrete reset map, and 
\( f, g \) define the continuous dynamics. Let $\x = [\q^\top, \dot{\q}^\top ]^\top \in \mathcal{TQ}$ where $\q$ is the generalized coordinate, the hybrid system's dynamics is given by:
\begin{align}
    \begin{cases} \dot{\x} &= f(\x) + g(\x) \boldsymbol{\tau}  \quad \x \in \mathcal{D} \setminus  \mathcal{G} , \\
    \x^+ &= \Delta( \x^-) \hspace{1.4cm} \x^-\in \mathcal{G}. \end{cases}
    \label{eq::HZ}
\end{align}

\subsection{Reduced Order Models for Underactuated Dynamics}
\begin{figure}
    \centering
    \includegraphics[width=\linewidth]{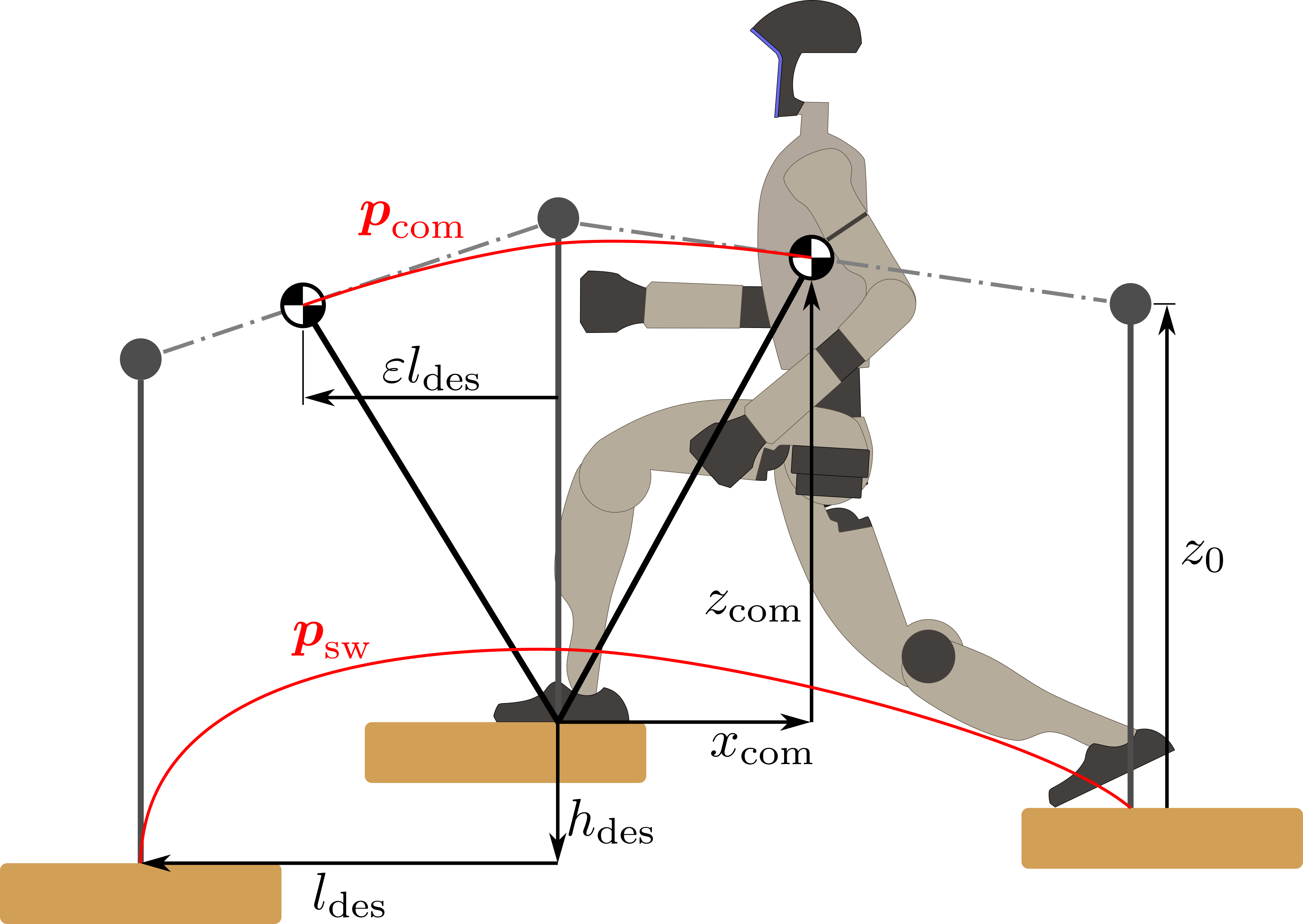} 
    \vspace{-5mm}
    \caption{A visual depiction of the LIP model used for gait synthesis, including the desired trajectories for the swing foot $\boldsymbol{p}_{\text{sw}}$ and CoM $\boldsymbol{p}_{\text{com}}$.}
    \label{fig:lip}
    \vspace{-3mm}
\end{figure}

Bipedal walking is inherently underactuated, even with ankle actuation. The horizontal CoM cannot be arbitrarily regulated, especially at moderate or high speeds. For example, when a human leans forward, it does not take long before the center of pressure reaches the edge of the foot, prompting a step to prevent falling. To capture this dynamic tendency, the passive-ankle LIP model is commonly used. As shown in Fig. \ref{fig:lip}, we model the varying height stepping stones as virtual slopes by connecting the centers of the consecutive stones. 

The LIP assumes the robot's CoM moves at constant height relative to the virtual slope along a massless leg with no ankle torque. Let $\mathbf{x} = \begin{bmatrix} p  &  L \end{bmatrix}^T$ represent the state vector, where $p$ is the CoM position relative to the stance foot and $L$ is the mass-normalized angular momentum about the stance foot. The sagittal-plane LIP dynamics can be expressed as:
\begin{align}\label{eq::LIPode}
\dot{\mathbf{x}} = 
\frac{d}{dt}
    \begin{bmatrix} p\\L\end{bmatrix} = \begin{bmatrix} 0 & \frac{1}{z_0}\\g & 0\end{bmatrix}\begin{bmatrix} p\\L\end{bmatrix}.
\end{align}
Here, $z_0$ is the constant vertical height of the CoM, and $g$ is the gravitational acceleration. This linear system captures the underactuated evolution of the CoM state during single support, where momentum is generated solely by gravity-induced torques about the stance foot.

\noindent \textbf{Orbital Energy and Step Viability: }
A key property of the LIP dynamics in \eqref{eq::LIPode} is the conservation of orbital energy, defined as:
\begin{align}
E = \left( \frac{L}{z_0} \right)^2 - \lambda^2 p^2, \quad \text{with } \lambda = \sqrt{\frac{g}{z_0}}.
\end{align}
The orbital energy $E$ partitions the state space into regions of viable motion: $E > 0$ corresponds to states with sufficient momentum to carry the CoM past midstance (forward walking), while $E < 0$ indicates bounded oscillations. Each orbital energy level defines a manifold of valid trajectories with varying boundary conditions. We leverage this property for stepping stones by selecting a target $E^*$ that ensures forward progression.

\noindent \textbf{Step Timing via Time-to-Impact: }Given the linear structure of \eqref{eq::LIPode}, the CoM evolution admits a closed-form solution:
\begin{align}
p(t) = \cosh(\lambda t)\,p_0 + \frac{1}{\lambda z_0} \sinh(\lambda t)\,L_0 ,
\end{align}
where $(p_0, L_0)$ is the initial state at the beginning of the step. 
This closed-form trajectory allows us to analytically compute the step duration $T$ required for the CoM to reach a target switching state $p(T) = \epsilon l_\text{des}$, where $l_\text{des}$ is the distance to the next stepping stone from the current stance foot, as depicted in Fig. \ref{fig:lip}. This calculation ensures precise timing regulation over non-uniform footholds, which has proven crucial in the literature \cite{khadiv_walking_2020,  li_gait-net-augmented_2025}.

\noindent \textbf{Impact and Angular Momentum Reset Map: }The robot’s hybrid dynamics involve discrete changes at foot impact. At impact, the swing foot makes contact and becomes the new stance foot. Assuming instantaneous, inelastic impact and no slipping, the angular momentum about the new contact point is reset as:
\begin{align}\label{eq:momentum_transfer}
L^+ = L^- + x_\text{sw} \dot{z}_\text{com}^- - z_\text{sw} \dot{p}^- , 
\end{align}
where $(x_\text{sw}, z_\text{sw})$ is the position of the swing foot before impact, and $\dot{p}^-, \dot{z}_\text{com}^-$ are the CoM velocities just prior to impact. Critically, this map reveals that for a fixed foothold $\boldsymbol{p}_\text{sw}$, the post-impact momentum $L^+$ can be actively regulated by modulating the pre-impact vertical velocity $\dot{z}_\text{com}^-$. We leverage this relationship to ensure the post-impact state always lies on the viable manifold defined by $E^*$.

\subsection{Nonlinear Output Dynamics and Control}
Let \( \boldsymbol{y} = h(\q) \in \mathbb{R}^{n_y} \) denote the set of controllable outputs (e.g., CoM position, torso orientation, swing foot pose).
For relative degree two outputs, the output dynamics satisfy
\begin{equation}
\ddot{\boldsymbol{y}} = L_f^2 h(\q,\dot{\q}) + L_g L_f h(\q,\dot{\q}) \boldsymbol{\tau},
\end{equation}
To ensure exponential convergence of outputs, we define a Control Lyapunov Function (CLF)
\begin{equation}
V(\boldsymbol{\eta}) = \frac{1}{2} \boldsymbol{\eta}^{\top} P \boldsymbol{\eta}, \quad 
\boldsymbol{\eta} = \begin{bmatrix} \boldsymbol{y}^a - \boldsymbol{y}^d \\ \dot{\boldsymbol{y}}^a - \dot{\boldsymbol{y}}^d \end{bmatrix},
\end{equation}
with \( P > 0 \) obtained from a continuous-time algebraic Riccati equation (CARE). The CLF decreasing condition 
\begin{align}
  \dot{V}(\boldsymbol{\eta}) \leq -c V(\boldsymbol{\eta}) 
\end{align}
is typically enforced through a quadratic program (QP) to compute the control input $\boldsymbol{\tau}$ that provides an exponential tracking condition on the output dynamics while satisfying the constrained dynamics and respecting the torque limit and contact constraints. 
In this work, CLF is directly embedded in the RL training as rewards as in \cite{li_clf-rl_2025}.

\subsection{Reinforcement Learning}

We formulate the task of sparse foothold traversal as a partially observable Markov decision process (POMDP), defined by the tuple $\mathcal{M} = (\mathcal{S}, \mathcal{A}, \mathcal{O}, \mathcal{T}, \mathcal{Z}, \mathcal{R}, \gamma)$, where $\mathcal{S}$ is the set of states, $\mathcal{A}$ is the set of actions, $\mathcal{O}$ is the set of observations from proprioception and perception, $\mathcal{T}(\boldsymbol{s}'|\boldsymbol{s},\boldsymbol{a})$ is the state transition probability,  $\mathcal{Z}(\boldsymbol{o}|\boldsymbol{s})$ is the observation likelihood, $\mathcal{R}(\boldsymbol{s},\boldsymbol{a})$ is the instantaneous reward function and $\gamma \in [0,1]$ is the discount factor. The agent seeks to maximize the expected return
\begin{equation}
J(\pi_\theta) = \mathbb{E}_{\pi \sim \pi_\theta} 
\left[ \sum_{t=0}^{T} \gamma^t r_t \right],
\end{equation}
where \( \pi_\theta(\boldsymbol{a}_t|\boldsymbol{s}_t) \) is the stochastic policy parameterized by \(\theta\).

\noindent \textbf{Proximal Policy Optimization (PPO): }
We use PPO~\cite{schulman_proximal_2017} to optimize both the teacher and student policies. PPO improves training stability by constraining the policy update through a clipped surrogate objective:
\begin{equation}
\mathcal{L}_{\mathrm{PPO}} = 
\mathbb{E}_t \left[
\min \left(
\rho_t(\theta) \hat{A}_t, 
\mathrm{clip}\big(\rho_t(\theta), 1-\epsilon, 1+\epsilon\big) \hat{A}_t
\right)
\right],
\end{equation}
where \( \rho_t(\theta) = \frac{\pi_\theta(\boldsymbol{a}_t|\boldsymbol{s}_t)}{\pi_{\theta_{\mathrm{old}}}(\boldsymbol{a}_t|\boldsymbol{s}_t)} \) is the importance sampling ratio and \( \hat{A}_t \) is the advantage estimate.

\noindent \textbf{Teacher-Student Distillation: }
To improve learning under partial observability, we adopt a teacher-student framework. The teacher's policy $\pi_T(\boldsymbol{a}_t | \boldsymbol{o}_t, \boldsymbol{o}_t^{\text{priv}})$ has access to privileged information such as ground-truth terrain and contact phase. During each environment step, both the teacher policy and the student policy $\pi_s(\boldsymbol{a}_t | \boldsymbol{o}_t)$ compute their respective actions. The student is then trained to match the teacher’s output using a per-step supervised behavior cloning loss:

\begin{equation}
\mathcal{L}_{\text{distill}} = \mathbb{E}_{t} \left[ \left\| \pi_s(\boldsymbol{o}_t) - \pi_T(\boldsymbol{o}_t, \boldsymbol{o}_t^{\text{priv}}) \right\|^2 \right].
\end{equation}
During distillation, the student policy minimizes this loss over on-policy rollouts.  
After distillation, the student is further fine-tuned using PPO to adapt to the partially observed setting and improve robustness under noise. This two-stage training procedure allows us to deploy a policy that retains the teacher's precision without relying on privileged inputs, and is especially useful when standard asymmetric actor-critic approaches fail to converge due to high information asymmetry.

\section{Model-Based Reference Generation}
\label{sec:model-based}

As shown in Fig. \ref{fig:arch_diagram}, our control approach integrates a model-based locomotion planner with a teacher-student learning framework. First, a reduced-order walking model is used to plan dynamically and environmentally consistent reference trajectories for each time step (foot placement, center-of-mass trajectory, and step timing). Second, a multi-stage teacher-student RL pipeline leverages these references. This sections detail the model-based planner and how it guides the RL training process.

\noindent \textbf{Step Time: }
Our planner builds upon the underactuated LIP model as in Sec. \ref{sec:preliminaries}. The step duration, $T_s$, is determined analytically based on the current post-impact state $\mathbf{x}^{+}$ and the required forward displacement $l_{des}$ to the upcoming stepping stone as shown in Fig. \ref{fig:lip}. $T_s$ is solved from the closed-form solution of the LIP dynamics, which specifies the time required for the horizontal CoM position to reach a target pre-impact location \cite{dai_bipedal_2022}, $x_{\text{com}}^{d} = \epsilon l_{des}$, where $\epsilon \in (0, 1)$. In this work, we used $\epsilon = 0.6$. This process dynamically updates step timing, rather than relying on interpolation among gait libraries or fixed timing assumptions.

\noindent \textbf{Orbital Energy Regulation using Vertical CoM Reference: } We regulate the post-impact orbital energy in the sagittal plane $E^+$ to a desired $E^* = 0.6$ through regulating the momentum $L_y$ at impact. In particular, we regulate the desired pre-impact vertical CoM velocity, $u_\text{des} = \dot{z}_\text{com}^-$ as in \cite{dai_bipedal_2022}. This target velocity acts as a discrete input to the hybrid system, calculated using the momentum transfer law derived from the underactuated robot dynamics as in Eq. \eqref{eq:momentum_transfer}. The desired vertical velocity $u_\text{des}$ is calculated based on the required change in angular momentum and the stone configuration:
\begin{align}
    u_\text{des} =  \frac{1}{l_\text{des}}(\hat{L}_\text{des} - \hat{L}_y^-(\x) + h_\text{des}\hat{\dot{x}}_{\text{com}}^-), \label{eq::dzcmf_estimation}
\end{align}
where $\hat{\cdot}$ is used to denote the estimation of a parameter using LIP dynamics and $\hat{L}_\text{des}$ corresponds to the desired orbital energy $E^*$ at post-impact. The estimation of pre-impact underactuated states $\hat{x}_\text{com}^-$ and $\hat{L}_y^-$ is performed using LIP solution given the time-to-impact. Other pre-impact states, including $z_{\text{com}}^{-}$ and $\hat{\dot{x}}_{\text{com}}^-$ are approximated by weighted averages of the desired pre-impact ones and the current ones. 

\noindent \textbf{CoM Translation Reference: }
For CoM motion, we replace the MPC-style planner used in prior model-based work \cite{dai_bipedal_2022} with closed-form cubic splines that provides smooth references inside the RL training loop. The splines are parameterized by the phasing variable $s = \frac{t}{T_s}\in [0,\, 1)$, ensuring time-normalized trajectories across different step durations. The coefficients are obtained by enforcing boundary conditions on position and velocity at the beginning and end of each step. The vertical spline boundaries are set to satisfy momentum regulation, while the sagittal spline connects the current state to a target end-of-step state derived analytically from the desired orbital energy $E^*$. To ensure lateral balance, the reference follows the analytical solution of the HLIP model \cite{xiong_3-d_2022}, directly mapping the planner's desired velocity and position to the training reference.

\noindent \textbf{Swing Foot Translation Reference: }
The swing-foot trajectory is generated by interpolating between the lift-off state and a target foothold using Bézier polynomials for all three dimensions. The target $x_\text{sw}$ and $z_\text{sw}$ are strictly determined by the geometry of the upcoming stepping stone. The lateral target $w_\text{des}$ is produced by the HLIP reference to regulate step width. The final trajectory evolves as:
\begin{align}
    {x,y^d_{\text{sw}}}(s) &= (1-b_h(s))x,y_{\text{sw}}(\x^+) +b_h(s)l,w_\text{des},\\
    {z^d_{\text{sw}}}(s) &= b_v(s),
\end{align}
where $b_h$ and $b_v$ are sets of Bézier polynomials \cite{grizzle_models_2014}. The coefficients of $b_h$ are $[0,0,0,\mathbf{1}_3]$, where $\mathbf{1}_N$ indicates a row vector of size $N$ with all elements being 1. The coefficients of $b_v$ are $[z_{\text{sw}}(\x^+), z^\text{max}_\text{sw} \mathbf{1}_3, h_{des},  h_{des}]$, where $z_{\text{sw}}^{\text{max}}$ determines the swing clearance and is computed from the maximum height of the previous, current, and upcoming stones with an added safety margin.

\noindent \textbf{Other References: } To ensure stable contact and natural posture, the reference orientation for both the pelvis and the swing foot is aligned with the local terrain heading. Simultaneously, the arm joints track a nominal, phase-synchronized reference trajectory to aid in angular momentum cancellation and generate naturalistic gait dynamics.

Collectively, this planner functions as a real-time reference generator, producing a full state trajectory at every time step. This includes the desired output $\boldsymbol{y}_d$, the actual output $\boldsymbol{y}_a$, and holonomic constraint information. The tracked outputs include CoM position and velocity, swing-foot position and velocity, pelvis and swing-foot orientation and angular velocity, and upper-body joint states. The holonomic constraints contain stance-foot position, velocity, orientation, and angular velocity, ensuring that ground-contact conditions remain consistent throughout single support. In the subsequent training phase, these trajectories are used throughout training to define CLF-based rewards, enforce CLF decreasing conditions, and incorporate holonomic and holonomic-velocity constraints within the RL objective \cite{li_clf-rl_2025}.

\section{Learning Environment Setup}
\label{sec:rlSetup}

To effectively leverage the model-based references described in \ref{sec:model-based}, the training framework operates hierarchically. The reduced-order model functions as a high-level stepping-stone planner, dictating the step timing $T_s$ and dynamic targets $u_\text{des}$. The RL policy then serves as a low-level, high-frequency controller that learns the residual full-body dynamics necessary to robustly track these targets on the high-DoF humanoid model. This fusion combines the strengths of both paradigms: the model-based planner generates references consistent with environmental constraints, while the RL policy handles unmodeled dynamics (e.g., friction, compliance) and minimizes tracking errors using the full robot actuation capabilities. This hybrid approach mitigates the brittleness of pure model-based control while significantly improving sample efficiency compared to model-free baselines as in \cite{wang_beamdojo_2025, he_attention-based_2025}, which often struggle to discover precise stepping strategies from scratch.

\subsection{Terrain Generation}
We construct a family of procedurally generated stepping-stone terrains with randomized geometric properties. A terrain instance is drawn from four terrain types: upstairs, downstairs, flat stepping stones, and height-varying stepping stones. These terrains differ in their sagittal spacing, vertical offsets, and the periodicity of the stone sequence.

We adopt a difficulty-indexed terrain sampling that incrementally expands the geometric variability of the stones. In general, at low difficulty, the stones are placed with small gaps with less variation and less vertical variation (if any). As difficulty increases, the allowable sagittal gap widens toward the maximum stone spacing; the stone height offsets expand symmetrically to introduce positive and negative elevation changes. 
Each episode consists of a procedurally generated terrain with parameters sampled from:
\begin{itemize}
    \item Gap distance: $[\,0.3, 0.3+0.4d \,]$m
    \item Height variation: $[\,-0.2d, 0.2d\,]$m
    \item Stone x: $[\,0.13, 0.3\,]$m
    \item Stone z: $[\,0.75, 1.25\,]$m
    \item Stair depth: $[\,0.2, 0.3\,]$m
    \item Absolute stair height: $[\,0., 0.2d\,]$m
\end{itemize}
where $d \in [0, 1]$ represents terrain difficulty and corresponds to 10 discrete terrain levels. These parameters govern the placement, shape, and elevation of each stone. In addition, the supporting platform under the stones is randomized within a bounded vertical range $[0, \, -1]$m. This produces a unified progression across all types of terrains, forcing the policy to learn generalizable stepping behaviors that remain stable under variations in gap width, vertical spread, and platform height. 

To enable stable learning on challenging foothold arrangements, we implement a curriculum that adjusts terrain difficulty automatically based on task performance. At the end of each episode, the environment evaluates whether the robot successfully traversed the terrain. A terrain level advances only after three consecutive successful traversals, preventing noisy or accidental successes from triggering progression. The curriculum is monotonic: once an environment moves to a higher difficulty level, it never regresses. Because higher levels sample from a wider parameter space that still includes easier configurations, the agent continues to experience a mix of difficulties while steadily shifting toward more challenging foothold patterns. This approach ensures smooth progression and encourages the emergence of robust, generalizable locomotion behaviors across the entire terrain family.

\subsection{Training Framework}

The CLF-based reward formulation is inherently phase-dependent, aligning with terrain structure and command sequences. However, this phase information is unavailable at deployment, preventing its direct use in the student's observation space. Furthermore, we observed that a standard asymmetric actor-critic setup—where only the critic has access to privileged information—failed to converge on the stepping stone task. The optimization landscape for such precise, sparse-foothold locomotion is too difficult for PPO to traverse from scratch; without a structured prior, the policy consistently collapses to a local optimum (e.g., standing still). To overcome this, we adopt a teacher-student distillation framework designed to provide the student policy with a robust initialization prior. The pipeline proceeds in three steps, outlined in Fig. \ref{fig:arch_diagram}:
\begin{enumerate} 
\item Teacher Training: A privileged teacher policy is first optimized to master basic locomotion (standing and walking) on nominal terrain. This phase leverages the full CLF-based reward structure and ground-truth phase information to generate dynamically consistent motion. 
\item Distillation (Warm Start): We distill this basic competence into the student policy using behavior cloning. This step removes the dependency on privileged information, effectively "warm-starting" the student with a stable, non-privileged walking prior. 
\item Curriculum Fine-tuning: Finally, the initialized student is trained via PPO on the full stepping-stone curriculum. Because the distillation phase positions the student in a favorable region of the optimization landscape, it can successfully adapt to the challenging height-varying stones and injected noise. \end{enumerate}
The student policy operates with deployable observations:
\begin{itemize}
    \item \textbf{Command Vector} $\boldsymbol{c}_t \in \mathbb{R}^d$: encodes task-specific information such as desired velocity.
    \item \textbf{Proprioception} $\boldsymbol{o}_t^{\text{proprio}} \in \mathbb{R}^{n}$: consists of base angular velocity $\boldsymbol{\omega}_t$, projected gravity vector $\boldsymbol{g}_t$, joint positions, and joint velocities.
    \item \textbf{Exteroception} $\boldsymbol{o}_t^{\text{extero}}$: a local 1m$\times$1m heightmap, with grid size 0.1m, capturing terrain geometry.
    \item \textbf{Last Action} $\boldsymbol{a}_{t-1} \in \R^{21}$
\end{itemize}
The actions $\boldsymbol{a}_t \in \R^{21}$ encode desired angles for the 12 lower body joints, waist yaw joint, as well as 8 upper body joints. A PD controller is used for the low-level tracking with desired joint velocities trivially zero. Additionally, the teacher policy has access to privileged inputs, including stance foot and phase-related CLF terms and targeted swing foot position given terrain information and current stance foot location. 

During teacher training and distillation, all observations are noise-free to ensure stable supervision. After distillation, the student policy is fine-tuned using reinforcement learning with deployable observations and injected noise to enhance generalization. The fine-tuning stage uses the same reward without access to teacher supervision or privileged inputs. This two-stage pipeline enables the student policy to first acquire structured behavior from a privileged teacher and then adapt to noisy and partial observations during fine-tuning for real-world deployment readiness.


\begin{figure}[!t]
    \centering
    \includegraphics[width=\linewidth]{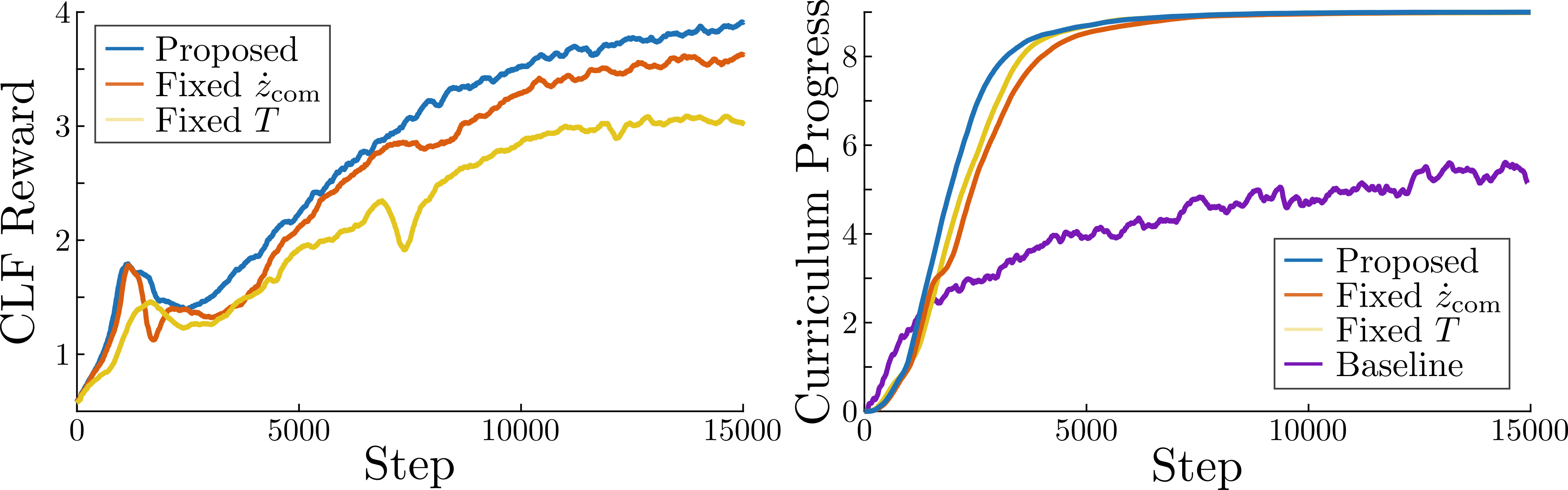} 
    \vspace{-5mm}
    \caption{(Left) The CLF reward curves during student policy training for the model-guided policies. The proposed method obtains better tracking than both the Fixed $\dot{z}_{\text{com}}$ and Fixed $T$ methods. (Right) A comparison between the model-guided policies and a naive policy; the naive policy is not able to handle the more difficult terrains.}
    \label{fig:training}
    \vspace{-3mm}
\end{figure}
\begin{figure}[!t]
    \centering
    \includegraphics[width=\linewidth]{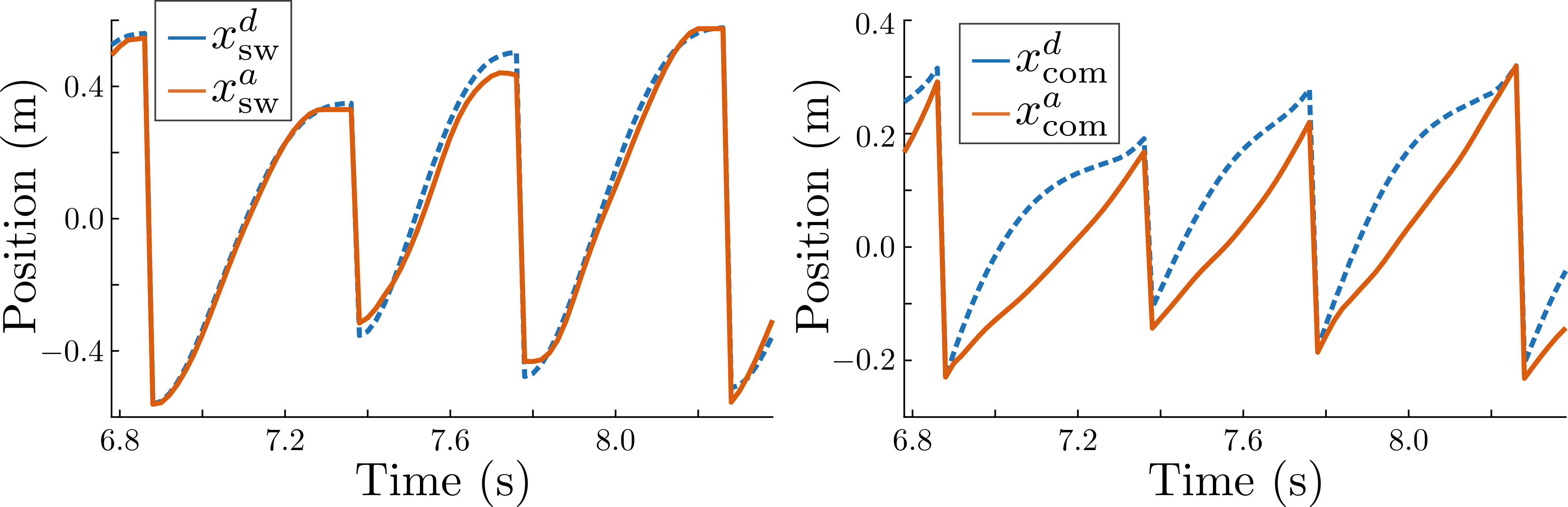} 
    \vspace{-5mm}
    \caption{The swing foot (left) and CoM (right) tracking in the $x$ direction. Precise foot tracking is critical to hit constrained footholds, and CoM tracking is particularly good around contact, improving performance on future stones.}
    \label{fig:tracking}
    \vspace{-3mm}
\end{figure}
\section{Results}
\label{sec:results}

All training is performed using a 21-DoF Unitree G1 humanoid in IsaacLab~\cite{mittal2025isaaclab} with IsaacSim \cite{isaacsim} physics engine. The robot is tasked with traversing randomized stepping-stone terrains and stair terrains. The policy outputs desired joint position targets at 50~Hz, which are tracked by a low-level PD controller. 

\noindent  \textbf{Hyperparameters:} We train using PPO with $4096$ parallel environments using an MLP network with $[512,256,128]$ hidden dimensions. Each policy is trained for $16$k iterations from the initial flat ground walking policy. Details for training parameters are provided in Table \ref{tab:training_configs} in the Appendix.

\noindent \textbf{Domain Randomization:} To facilitate sim-to-real transfer, we apply standard domain randomization to the robot's inertial properties, like CoM position and base mass.  Contact dynamics are randomized with friction coefficients sampled uniformly between $0.4$ and $1.0$. Noise is injected into observations, including the elevation map. To enhance disturbance rejection, we apply random velocity perturbations to the base during training, with linear velocities up to $0.5$~m/s and angular velocities up to $0.4$~rad/s.

\noindent  \textbf{Baseline Construction:}
As prior stepping-stone RL frameworks do not release implementation details, we implement our own direct end-to-end RL baseline. To ensure a fair comparison, the baseline uses the same student-level observation space, identical PPO hyperparameters, and the reward structure provided by the Unitree RL Lab~\cite{unitree_rllab}.  
When trained directly on the full stepping-stone environment, the policy immediately collapses to standing and makes no learning progress. To give the baseline the best possible chance, we adopt a two-stage curriculum: the policy is first pre-trained for 2k iterations on nominal flat-ground walking, and then fine-tuned exclusively on flat stepping stones. This modified curriculum allows the baseline to achieve non-zero performance on simple terrains for comparison.

\noindent  \textbf{Ablation Construction:} In addition to the end-to-end RL baseline, we conduct two ablations to isolate the contribution of each component in the model-based reference: \emph{fixed step timing} ($T = 0.4\text{ s}$) and \emph{no momentum shaping} ($u_{\text{des}} = 0$).

\subsection{Quantitative Results}
\begin{table}[t]
\centering
\caption{Success rate across terrain conditions for baseline, ablations, and the proposed method.}
\label{tab:success_rate}
\begin{tabular}{lcccc}
\toprule
\textbf{Terrain} & \textbf{Baseline} & \textbf{Fixed $\dot z_{\mathrm{com}}$} & \textbf{Fixed $T$} & \textbf{Proposed} \\
\midrule
Flat (mid difficulty)          & 0.92896 & \textbf{1} & \textbf{1} & \textbf{1} \\
Flat (high difficulty)         & 0.55762 & 0.99854 & \textbf{1} & \textbf{1} \\
Height-varying (mid)           & 0       & 0.99194 & 0.99902 & \textbf{1} \\
Height-varying (high)          & 0       & 0.94116 & 0.87915 & \textbf{0.97217} \\
\bottomrule
\end{tabular}
\vspace{-3mm}
\end{table}

\begin{figure*}[!t]
\vspace{0.2cm}
    \centering
    \includegraphics[clip, trim=0cm 14.7cm 0cm 0cm, width=2.05\columnwidth]{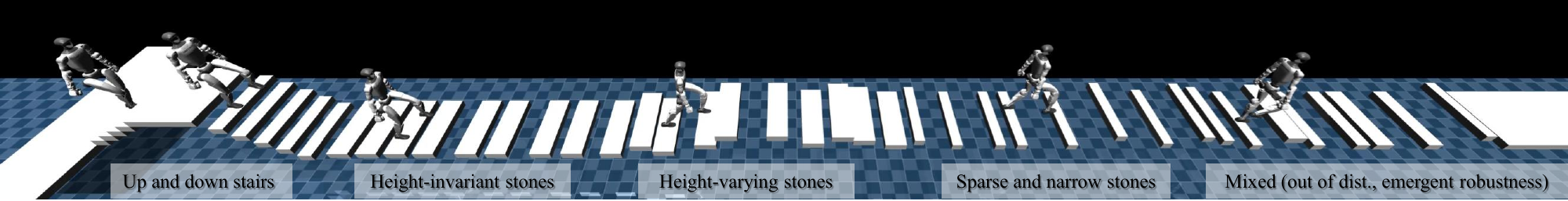}
     \caption{Snapshots of humanoid locomotion over a long stepping-stone course in MuJoCo simulation.}
        \label{fig:longsim}
        \vspace{-0.4cm}
\end{figure*}

\noindent  \textbf{Comparison with End-to-end Baseline: }
Across all experiments, the proposed model-guided RL policy and its two ablations demonstrate consistent and stable learning behavior, rapidly converging to high performance. In contrast, the end-to-end baseline improves only on flat stepping stones. While it learns to traverse flat stones, it saturates at a much lower curriculum level and ultimately converges to an asymmetric period-2 gait, consistent with findings in~\cite{wang_beamdojo_2025}.  

This contrast highlights a key limitation of pure RL: without model-based structure, the policy struggles to discover the precise, height-aware foot placements necessary for stepping-stone locomotion. Our model-guided approach leverages the RoM-based planner references to efficiently acquire feasible stepping strategies and refine them throughout training. Notably, as reported in Table \ref{tab:success_rate}, it attains 100\% success on challenging flat stepping stones, compared to only 56\% for the end-to-end RL baseline. The success rate is evaluated over 4096 environments.

\noindent  \textbf{Ablation Analysis: }
To understand the contribution of each component in the model-guided RL pipeline, we conduct ablation studies that remove either (i) step timing adaptation or (ii) vertical COM momentum regulation. The CLF-reward traces (Fig.~\ref{fig:training}, left) show that all model-guided variants learn rapidly and stably, exhibiting a consistent upward trend throughout training. The full proposed method achieves the highest CLF reward, reflecting better satisfaction of the model-based stability objective. Removing vertical COM regulation lowers performance, while fixing the step timing produces more pronounced degradation, with slower improvement and reduced steady-state reward. Interestingly, RL remains robust even under these ablations: despite the missing structure, the policies partially compensate by rediscovering surrogate strategies—either by implicitly regulating momentum through other reward channels or by learning to “survive” with less structured behavior.

Across evaluation terrains, all model-guided approaches achieve high success rates in simpler settings, including both difficulty levels of flat stepping stones and the mid-difficulty height-varying terrain. The advantage of the proposed method becomes particularly evident on the most challenging height-varying task, where it outperforms both ablations (Table~\ref{tab:success_rate}).

\subsection{Simulation Evaluation}
The high success rates on challenging terrain can be attributed to the policy's ability to track the model-based references, as illustrated in Fig.~\ref{fig:tracking}. The proposed method exhibits precise tracking of the swing foot and CoM trajectories in the sagittal direction. To assess the generality and robustness of the learned locomotion strategies before transfer to hardware, we perform a sim-to-sim transfer study in which all final policies are deployed in a separate MuJoCo environment with independently parameterized dynamics, contact models, and actuator responses. As shown in Fig.~\ref{fig:longsim}, the proposed method successfully traverses an extended, mixed-terrain course that includes up-and-down stairs, height-invariant stepping stones, height-varying stones with large vertical offsets, sparse and narrow footholds, and an out-of-distribution mixed sequence designed to probe emergent robustness. These results highlight that the reduced-order structure embedded in the proposed approach not only accelerates learning but also significantly enhances transferability across simulation platforms.

\begin{figure}[!t]
     \centering
    \includegraphics[clip, trim=0cm 9.5cm 0cm 0cm, width=1\columnwidth]{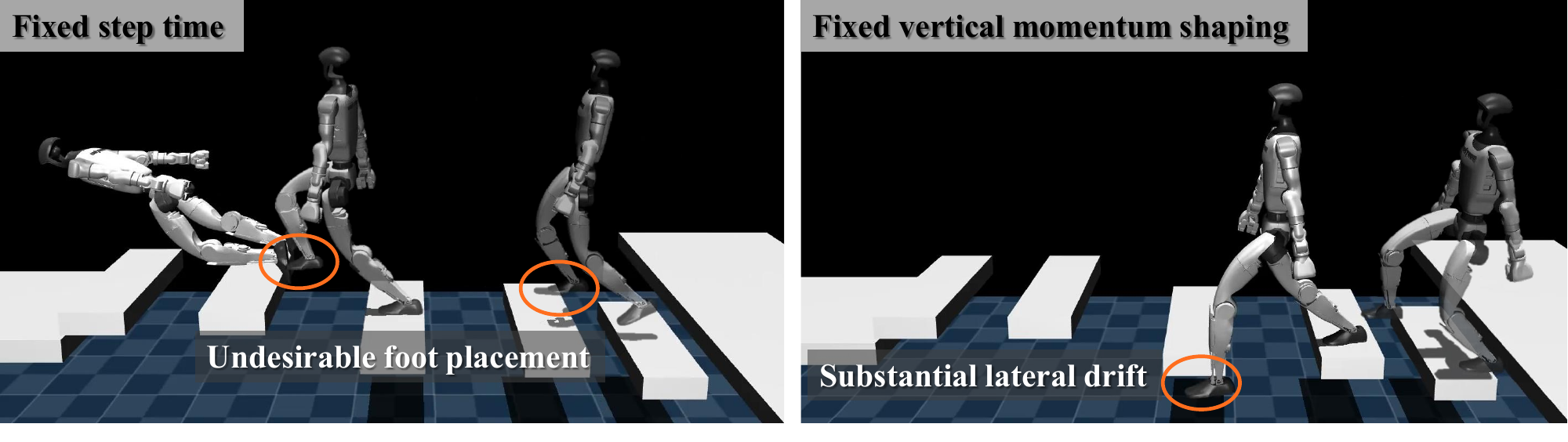}
     \caption{Snapshots of sim-to-sim ablation trials.}
        \label{fig:ablationsim}
        \vspace{-0.3cm}
\end{figure}

Moreover, the ablated policies exhibit reduced sim-to-sim transfer reliability, as reflected in gait irregularities and inconsistent stepping patterns observed in the gait tiles (Fig.~\ref{fig:ablationsim}). This sensitivity underscores the stabilizing role of both step timing adaptation and vertical COM regulation in promoting consistent, transferable locomotion strategies. For instance, the fixed $T$ case exhibits inaccurate robot foot placements and undesirable flight phases; In the fixed $\dot z_\textrm{com}^-$ case, the robot has significant $y$-direction drift.

Additionally, the resulting policy from the proposed method is robust to withstand external perturbations during stepping-stone traversal, such as unknown pushes to the robot body in the range of $\pm 100$Nm for a 0.2s duration. We refer readers to the supplementary videos for more details.

\subsection{Real-World Transfer}

\begin{figure}[!t]
     \centering
    \includegraphics[clip, trim=0cm 9.5cm 0cm 0.1cm, width=1\columnwidth]{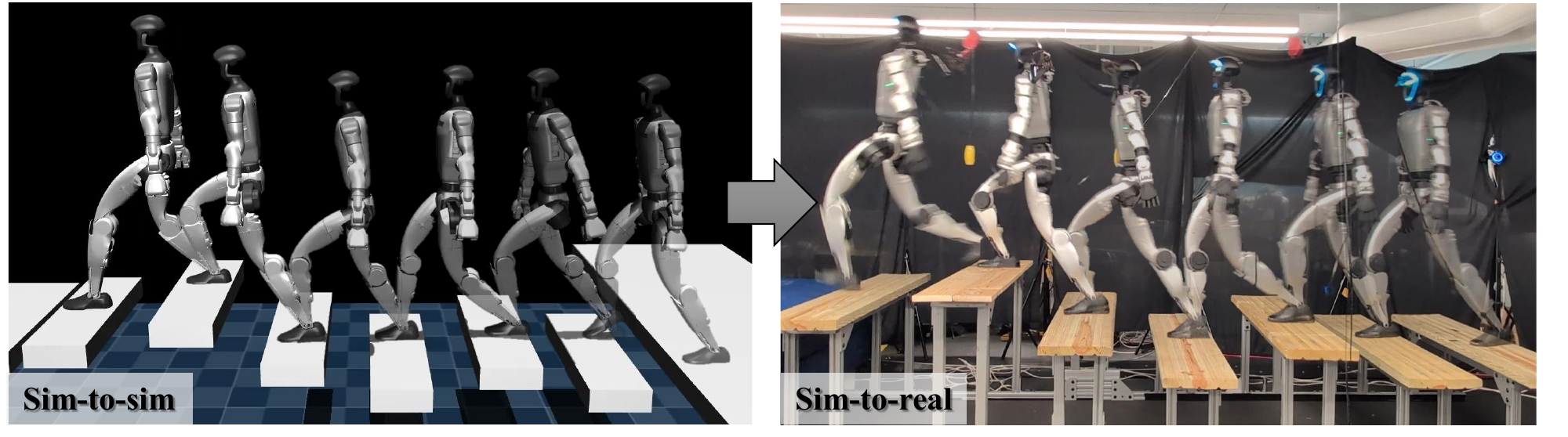}
     \caption{Comparison of sim-to-sim and sim-to-real snapshots.}
        \label{fig:sim2real}
        \vspace{-0.3cm}
\end{figure}




To evaluate our learned policy on hardware, we adopt a deployment pipeline similar to the GR-1 system described in~\cite{he_attention-based_2025}. The robot’s pose is tracked using a Qualisys motion capture system, and elevation maps are constructed by ray-casting onto a pre-designed terrain mesh that matches the physical testbed. Figure~\ref{fig:hero} shows gait tiles extracted from hardware rollouts of the G1 robot. The proposed controller produces a stable and repeatable stepping pattern with consistent swing trajectories and smooth torso motion, closely matching the behavior observed in simulation, as seen in Fig.~\ref{fig:sim2real}. The tiles illustrate that the learned policy maintains foot placement accuracy and gait symmetry despite unmodeled dynamics, actuator latency, and real-world disturbances. These results confirm that the model-guided structure enables reliable hardware deployment without additional tuning.

Beyond standard evaluations, we challenged the policy with a mixed stepping-stone course featuring irregular stone depths, which is a configuration absent from the training curriculum. Despite this distribution shift, the policy generalized successfully (Fig.~\ref{fig:hero}, top). This suggests that the model-guided references prevent the policy from overfitting to fixed geometric templates, enabling zero-shot generalization to unstructured terrain topologies.


\section{Conclusion}
In this work, we presented a model-guided RL framework that bridges the gap between the precision of model-based planning and the robustness of data-driven control for bipedal locomotion on constrained terrains. By integrating a reduced-order stepping planner directly into the training loop, we generated dynamically consistent references to guide the policy toward feasible motion manifolds without over-constraining the learning process. 
The successful zero-shot transfer to both a disparate physics simulator and physical hardware highlights the framework's effectiveness and robustness, offering a promising direction for deploying agile humanoids in complex, unstructured environments.



\ifCLASSOPTIONcaptionsoff
  \newpage
\fi



%

\bibliographystyle{IEEEtran}
\balance

\bibliography{references,misc}


%

\appendix

\vspace{-5mm}
\begin{table}[h]
\centering
\caption{Summary of training configurations used for PPO training, PPO fine-tuning, and student--teacher distillation.}
\label{tab:training_configs}
\begin{tabular}{lccc}
\toprule
\textbf{Parameter} & \textbf{PPO Train} & \textbf{PPO Finetune} & \textbf{Distillation} \\
\midrule
Steps per environment      & 24             & 24             & 120 \\
Init noise std             & 1.0            & 1.0            & 0.1 \\
Learning rate              & $1\!\times\!10^{-3}$ 
                           & $1\!\times\!10^{-4}$ 
                           & $5\!\times\!10^{-4}$ \\
Entropy coefficient        & 0.008          & 0.003          & --- \\
Clip parameter             & 0.2            & 0.1            & --- \\
 Learning epochs          & 5              & 5              & 10 \\
Grad norm max              & 1.0            & 1.0            & 2.0 \\
Loss type                  & clipped PPO    & clipped PPO    & Huber \\
Distillation gradient len  & ---            & ---            & 2 \\
\bottomrule
\end{tabular}
\end{table}

\end{document}